\documentclass[conference]{IEEEtran}
\IEEEoverridecommandlockouts
\usepackage{cite}
\usepackage{physics}
\usepackage{amsmath}
\usepackage{tikz}
\usepackage{mathdots}
\usepackage{yhmath}
\usepackage{multicol}
\usepackage{cancel}
\usepackage{color}
\usepackage{siunitx}
\usepackage{array}
\usepackage{multirow}
\usepackage{amssymb}
\usepackage{gensymb}
\usepackage{tabularx}
\usepackage{booktabs}
\usetikzlibrary{fadings}
\usetikzlibrary{patterns}
\usetikzlibrary{shadows.blur}
\usetikzlibrary{shapes}

\usepackage{amsmath,amssymb,amsfonts}
\usepackage{algorithmic}
\usepackage{graphicx}
\usepackage{textcomp}
\usepackage{hyperref}
\usepackage{xcolor}
\def\BibTeX{{\rm B\kern-.05em{\sc i\kern-.025em b}\kern-.08em
    T\kern-.1667em\lower.7ex\hbox{E}\kern-.125emX}}
\begin{document}

\title{CBIR using Pre-Trained Neural Networks \\
}

\author{\IEEEauthorblockN{Agnel Lazar Alappat}
\IEEEauthorblockA{\textit{Electronics and Communication Engineering} \\
\textit{Indian Institute of Information Technology, Nagpur}\\
Nagpur, India \\
pimpalkhutevarad@gmail.com}
\and
\IEEEauthorblockN{Ambarish Chandurkar}
\IEEEauthorblockA{\textit{Electronics and Communication Engineering} \\
\textit{Indian Institute of Information Technology, Nagpur}\\
Nagpur, India \\
@gmail.com}
\and
\IEEEauthorblockN{Sagar Suman}
\IEEEauthorblockB{\textit{Electronics and Communication Engineering} \\
\textit{Indian Institute of Information Technology, Nagpur}\\
Nagpur, India \\
@gamil.com}
\and
\IEEEauthorblockN{Prajwal Nakhate}
\IEEEauthorblockA{\textit{Electronics and Communication Engineering} \\
\textit{Indian Institute of Information Technology, Nagpur}\\
Nagpur, India \\
ambarishchdk@gmail.com}
\and
\IEEEauthorblockN{Varad Pimpalkhute}
\IEEEauthorblockA{\textit{Electronics and Communication Engineering} \\
\textit{Indian Institute of Information Technology, Nagpur}\\
Nagpur, India \\
pimpalkhutevarad@gmail.com}
\and
\IEEEauthorblockN{Tapan Jain}
\IEEEauthorblockA{\textit{Electronics and Communication Engineering} \\
\textit{Indian Institute of Information Technology, Nagpur}\\
Nagpur, India \\
@gmail.com}
}

\author{
    \IEEEauthorblockN{Agnel Lazar Alappat\IEEEauthorrefmark{1}, Prajwal Nakhate\IEEEauthorrefmark{1}, Sagar Suman\IEEEauthorrefmark{1}, Ambarish Chandurkar\IEEEauthorrefmark{1}, Varad Pimpalkhute\textsuperscript{a,}\thanks{\textsuperscript{\textbf{a}}\textbf{ Corresponding Author}}\IEEEauthorrefmark{1}, Tapan Jain\IEEEauthorrefmark{1}}\\
    \IEEEauthorblockA{\IEEEauthorrefmark{1}Electronics and Communication Engineering, Indian Institute of Information Technology, Nagpur (IIITN), India.
    \\}

    \thanks{\textbf{Email Addresses:} \href{mailto:agnel0033@gmail.com}{agnel0033@gmail.com} (Agnel Lazar Alappat), \href{mailto:prajwalnakhate@gmail.com}{prajwalnakhate@gmail.com} (Prajwal Nakhate), \href{mailto:sagarsuman52@gmail.com}{sagarsuman52@gmail.com} (Sagar Suman), \href{mailto:ambarishchdk@gmail.com}{ambarishchdk@gmail.com} (Ambarish Chandurkar), \href{mailto:pimpalkhutevarad@gmail.com}{pimpalkhutevarad@gmail.com} (Varad Pimpalkhute), \href{mailto:tapankumarjain@gmail.com}{tapankumarjain@gmail.com} (Tapan Jain)}
}

\maketitle

\textbf{\emph{Abstract - }Much of the recent research work in image retrieval, has been focused around using Neural Networks as the core component. Many of the papers in other domain have shown that training multiple models, and then combining their outcomes, provide good results. This is since, a single Neural Network model, may not extract sufficient information from the input. In this paper, we aim to follow a different approach. Instead of the using a single model, we use a pretrained Inception V3 model, and extract activation of its last fully connected layer, which forms a low dimensional representation of the image. This feature matrix, is then divided into branches and separate feature extraction is done for each branch, to obtain multiple features flattened into a vector. Such individual vectors are then combined, to get a single combined feature. We make use of CUB200-2011 Dataset, which comprises of 200 birds classes to train the model on. We achieved a training accuracy of 99.46\% and validation accuracy of 84.56\% for the same. On further use of 3 branched global descriptors, we improve the validation accuracy to 88.89\%. For this, we made use of MS-RMAC feature extraction method.}

\vspace{0.4cm}

\textbf{Keywords: CBIR; Deep Learning; MS-RMAC; Inceptionv3; Global Descriptors}
\section{Introduction}
Artificial Neural Networks(ANN) have made revolutionary breakthroughs in the area of computer vision. Image processing has been quite old field now and ANN have proven their might in various research areas ranging from \cite{lecun1990handwritten} recognising handwritten digits for zip codes , \cite{viola2001rapid}Face Detection which comes under the realm of Object Detection, Image Classfication with best examples like \cite{krizhevsky2012imagenet} AlexNet used in ImageNet\cite{imagenet_cvpr09} competition by krizhevsky et al. and varios other fields.\\

Content Based Image Retrieval (CBIR) is another such blooming field in the area of computer vision. CBIR refers to the act of picking out a number of images from a database, based on what's present ``inside" the query image. In CBIR, the general steps\cite{SIFTmeetsCNN} are to proceed by extracting some features or descriptors from the database and query images. Finally, both of the calculated features are compared, by taking their differences using some metric like euclidean distance, l2 norm, etc. However, feature engineering requires domain expertise in the field of Image Processing.\\

While the logic behind any CBIR algorithm is similar, the techniques have varied greatly\cite{SIFTmeetsCNN}. With the advent of ANN, much of the reserach in this decade has been  focused on using them for the purpose of Feature Extraction. This quite natural since ANN are termed as \cite{csaji2001approximation}``Universal Approximators". In this paper we used Pre-trained CNN (Name) for the purpose of extracting the descriptors of the image and apply own CNN structure on those descriptors in different branches and then concatenate them to form composite descriptor for the image input. \\

\begin{figure*}
    \centering
    \includegraphics[width = \linewidth
    ]{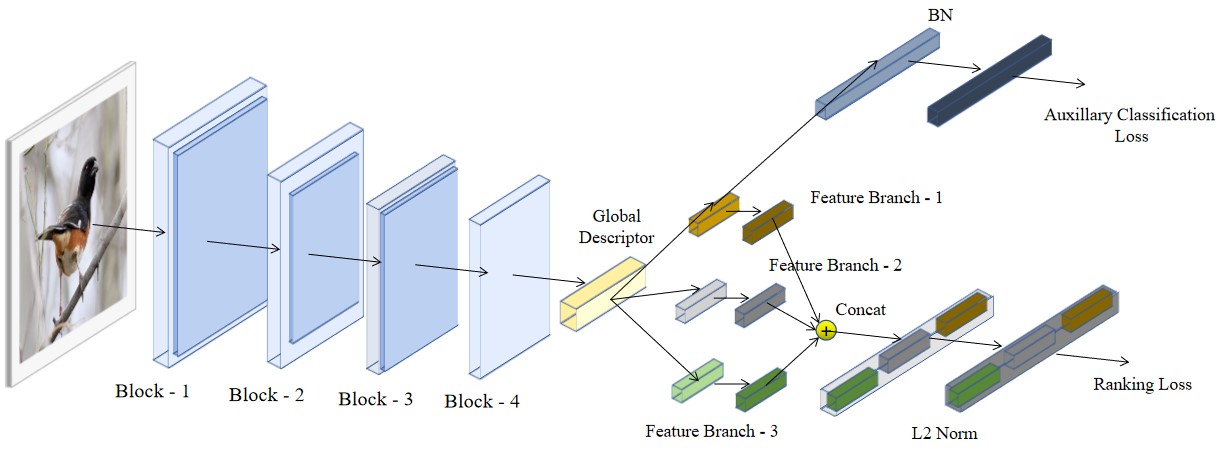}
    \caption{Proposed Methodology for Precise Image Retrieval}
    \label{fig:cgd}
\end{figure*}

In this paper, we aim to focus on how to get an ensemble-like effect by exploiting use of diverse global descriptors without the explicit need for multi-tasking and control of diversity amongst learners. Our contribution is threefold. 
\begin{itemize}
    \item[(1)] Global Descriptors have proven to be quite useful, thus we propose a 3 way branched combination of multiple global descriptors (CGD), that combines multiple global descriptors and retrieved using MS-RMAC feature extraction method. It achieves an ensemble-like effect without any explicit ensemble model or diversity control over each global descriptor~\cite{jun2019combination}. Furthermore, due to the proposed framework being adaptable to changes, we ensure that the architecture is able to generalize to new tasks.
    \item[(2)] We investigate the effectiveness of combining various multiple global descriptors and by experimentation figuring out how many branches result in the desired result. Our extensive experiments demonstrate that using combined descriptor outperforms a single global descriptor since it can use different types of feature properties ~\cite{jun2019combination}.
    \item[(3)] The proposed framework achieves promising results on datasets aside from CUB200-2011 (CUB200), CARS196 and SOP, further proving that our algorithm can be generalized to new tasks.
\end{itemize}

The paper is organized as follows: Section II gives a brief introduction to the past work on this research problem. Section III discusses the basics of pre-trained Inception v3 model, branching of global descriptors and retrieval strategy. Section IV presents the experiments performed and a brief methodology for the same. Benchmarking results and discussion are included in Section V and the paper is concluded in Section VI.

\section{Literature Survey}

The field of CBIR, has been researched with many techniques. As mentioned in \cite{895972}, CBIR in its early days, demanded domain knowledge in Image Processing and algorithms involved extraction of information about local shapes, textures, object features in the image. However, the usage of traditional methods, beagn decreasing thereafter.In 2004, Lowe et al\cite{lowe1999object}, published a paper on "Scale Invariant Feature Transform", which became the de facto standard. Much of the papers published later were based on SIFT based descriptors.\\

One of the examples, of such a method is Bag of Words (BoW) model\cite{Csurka2004VisualCW}.  It describes the method being robust to background clutters.It also compares the algorithm with alternative approaches implemented using Machine Learning algorithms like SVM and Naive Bayes. Some other examples, include \cite{perronnin2010improving} which uses Bag of Visual Words and a SVM classifier to perform high accuracy image classification. \cite{jegou2008hamming} uses  Hamming embedding (HE) and weak geometric consistency constraints (WGC) for improving the representation of the descriptors.\\

However, with the advent of Neural Networks, much of the CBIR research is focused on using its different implementations. The groundbreaking success of AlexNet \cite{krizhevsky2012imagenet}, in ImageNet\cite{imagenet_cvpr09} database classification, proved the potential of CNN. Since then, many papers were published utilising the CNN for image retrieval purpose too.\cite{yue2015exploiting}Achieved state-of-the-art retrieval results using low dimensional representations on two of the instance image retrieval datasets.It provides guidance for transferring deep networks trained on image classification to image retrieval task and then encode the features into single unique vector using VLAD\cite{arandjelovic2013all}. Another Paper \cite{ren2015faster}, achieves state-of-the-art object detection accuracy on PASCAL VOC 2007 (73.2\%mAP) and 2012 (70.4\% mAP) datasets using 300 proposals per image. This paper computes region proposals using CNN, and thus implements a novel Region Proposal Network. Both of the above papers justify the fact that CNN can be used extract generic features. In \cite{sharif2014cnn}, they use features extracted from the OverFeat\cite{sermanet2013overfeat} network as a generic image representation. The paper implements it for a range of recognition tasks of object image classification, scene recognition, fine grained recognition, attribute detection and image retrieval.\\

\begin{figure*}
    \centering
    \includegraphics[width = \linewidth]{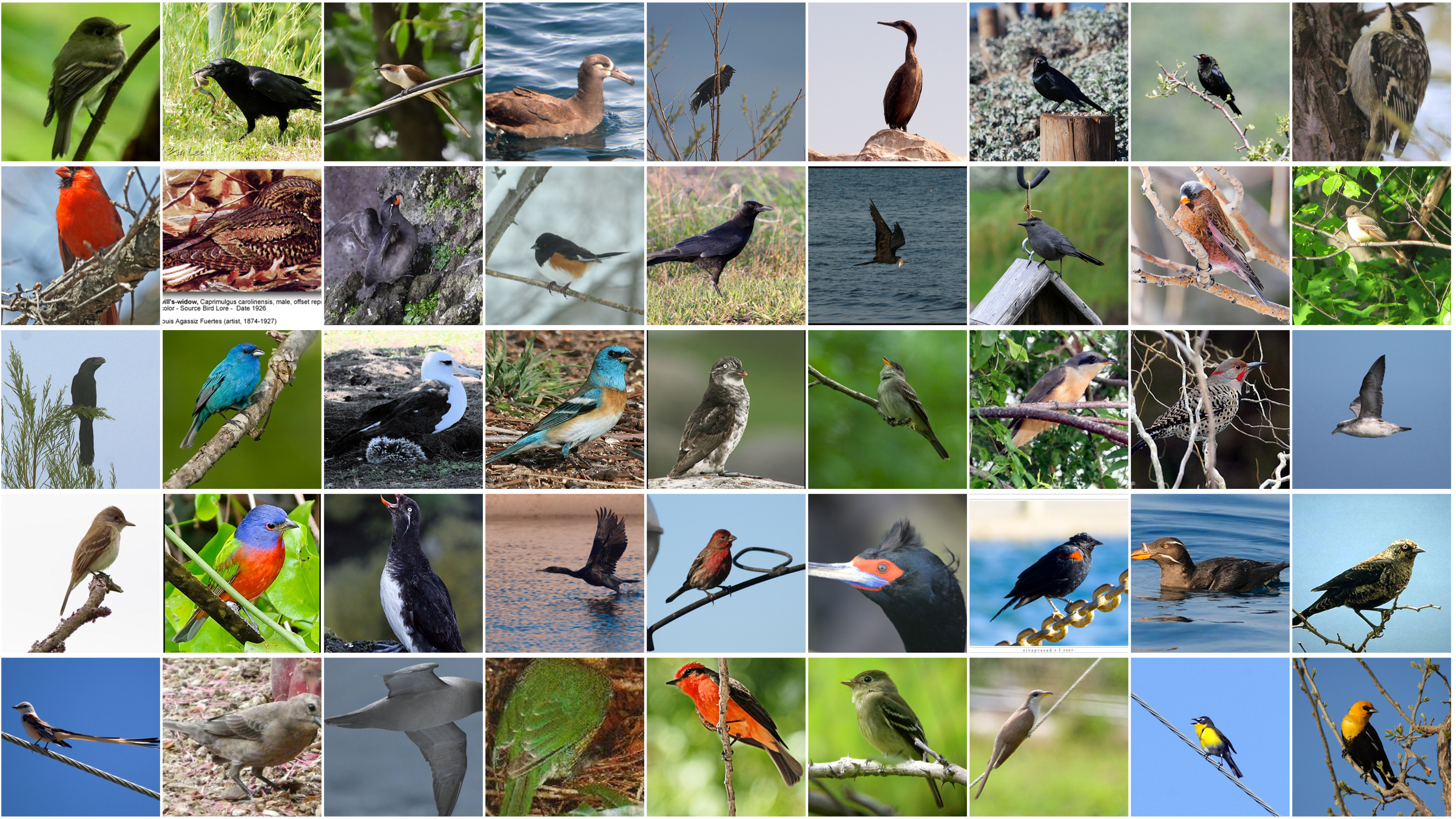}
    \caption{Some sample images from database CUB-200-2011}
    \label{fig:sample_images}
\end{figure*}

In \cite{xuan2018deep},the authors have proposed Deep Randomized Ensembles for Metric Learning (DREML), for creating an ensemble of diverse embedding functions.They illustrated the effect of different parameter choices relating to the embedding size and number of embeddings within the ensemble and demonstrated improvement over the state of the art for retrieval tasks for the CUB-200-2011 , Cars-196 , In-Shop Clothes Retrieval and VehicleID  datasets.
The authors have argued that ensemble based approaches outperform the non-ensemble approaches by a dramatic margin on all four datasets. In \cite{roth2020pads}, authors present a learned adaptive triplet sampling strategy using Reinforcement Learning.They provide a method for learning a novel triplet sampling strategy which is able to effectively support the learning process of a Deep Metric Learning (DML) model at every stage of training.They optimize a teacher network to adjust the negative sampling distribution to the ongoing training state of a DML model.By training the teacher to directly improve the evaluation metric on a held-back validation set, the resulting training signal optimally facilitates DML learning.  

In \cite{Lin_2015_ICCV}, the authors have proposed bi-linear CNN models, which consists of two feature extractors whose outputs are multiplied using outer product at each location of the image and pooled to obtain an image descriptor, which is called as `Bi linear Combination' of feature vectors. They report an accuracy of 84.1\% on the CUB201 - 2011 Dataset. Based upon this \cite{Lin_2015_ICCV} approach, \cite{srinivas2017deep} has proposed on-line dictionary learning (ODL) algorithm where the principle of sparsity is integrated into classification. It explains that the features learnt by BCCN are quite large and ODL represents it sparsely.They claim an accuracy of 84.6\%. Authors of \cite{han2018attribute} have proposed Attribute-Aware Attention Model ($A^{3}M$), which can learn both local and global image attributes simultaneously, in an end to end manner. They use CNN for feature extraction and then apply attention models on them.Using InceptionBN as their feature extractor, a Recall@1 of 61.2 is mentioned.\\

Papers like \cite{lin2018regional} have opted to use pretrained convolutional neural networks like VGG - 16 \cite{Simonyan15}, ResNet, ResNeXt, SEnet in order to use their potential for making low dimensional representation of the images. They use kNN for finding the nearest images to the query image. In this paper, we implement a similar strategy, utilising InceptionV3 \cite{szegedy2016rethinking} pre-trained CNN.

\section{Proposed Methodology}

The proposed architecture that we present is an unembellished and precise framework which is generated by concatenating multiple global descriptors in a single vector grid. Thus, we also call this Combined Multiple Global Descriptors or CGD in short. \\

Our proposed architecture is depicted in Figure \ref{fig:cgd}. The proposed architecture consists of a CNN based neural network and classes which extract particular features from input vector which help in computing loss. This loss accordingly trains the global descriptors and thus, what we get is a fine-tuned architecture which is able to effectively retrieve image belonging to classes on which the model has been trained on.

\begin{figure}[ht!]
    \centering
    \includegraphics[width = \linewidth]{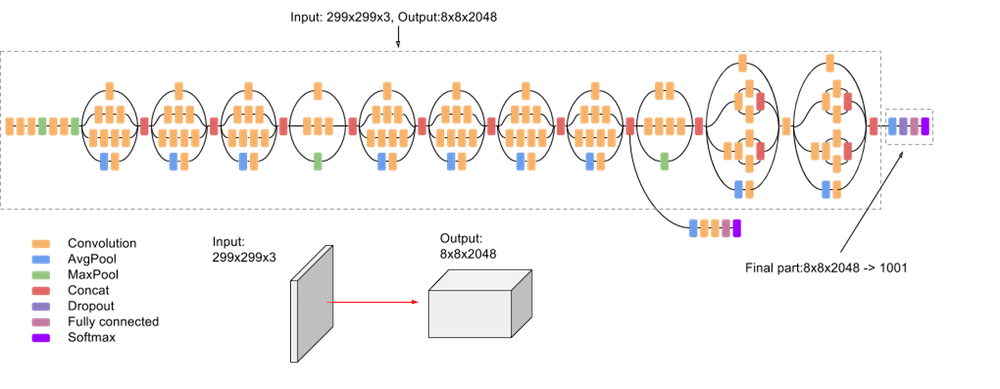}
    \caption{Inception v3 Architecture ~\cite{google:2020}}
    \label{fig:inceptv3}
\end{figure}

\subsection{The Base Pre-Trained Model}
As discussed in the last part of Literature Survey, \cite{jun2019combination}, we come to understand that a novel ResNet50 architecture has been trained as a base model for the analysis. We have provided a novel method in which a pre trained base model, InceptionV3 \cite{szegedy2016rethinking} is used. We observed, the same which we will discussing in Section V, Inception v3 gives comparatively better results when compared to ResNet50 architecture. The final layers are replaced with different branches of Dense layers. The output of these branches in concatenated to get the final Global descriptors. This entire model can be trained in an end-to-end fashion and gives us the ensemble effect without the need for running different CNN's for each descriptor. The authors of the paper have not used any diversity control \cite{kalantidis2016cross} \cite{oh2016deep} during the training of the model or branches. One interesting aspect of Inception v3 is there can be a lot of improvement on exploiting sub-branches. Thus, the number of branches will be affected by the number of parameters in a given hidden unit.\\

\subsection{The Branching}
The effectiveness of the aforementioned approach depends on the ensemble effect contributed by the branches which in turn depends on the diversity of the descriptors contributed by each branch. So in-order to ensure that each branch learns and outputs descriptors based on different features, we train each branch individually first. Each individual branch is trained until it achieves considerable accuracy itself. Each branch can be trained differently to learn different features. For example, a branch can trained as mentioned in \cite{opitz2017bier}, another branch can be trained as mentioned in \cite{GoogleRetreivalData}, etc. We append these branches to the base model and append an additional Dense layer which takes 
as input the concatenated descriptors of all these branches and outputs the final Global descriptor. Now, we discuss about the individual descriptors\\

\def\arraystretch{2}%
\begin{table*}[ht!]
    \centering
    \caption{Comparison of accuracy before branching and after branching}
    \begin{tabular}{c|c|c|c|c|c}
    \hline
    Layers Freezed for Training & Accuracy before branching & \multicolumn{4}{c}{Accuracy after branching} \\
    \cline{3-6}
    & & Branches = 2 & Branches = 3 & Branches = 4 & Branches = 5 \\
    \hline
    170 & 98.89 \% & 73.61 \% & 78.14 \% & 79.99 \% & 74.04 \%\\
    \hline
    \textbf{200} & \textbf{99.46 \%} & \textbf{87.63 \%} & \textbf{91.24 \%} & \textbf{88.08 \%} & \textbf{87.68 \%} \\
    \hline
    \end{tabular}
    \label{tab:no_branching}
\end{table*}

\subsubsection{\textbf{MS-RMAC Feature \cite{li2017ms}}}
MS-RMAC Stands for `Multi-Scale Regional Maximum
Activation of Convolutions for Image Retrieval'. As described in \cite{li2017ms}, it is the feature matrix representing the Input Image, made by concatenation of RMAC (Regional Maximum
Activation of Convolution) \cite{tolias2015particular} feature vectors, calculated by max pooling the $L^{th}$ convolutional Layer output, over specific regions $R_{i}$.\\
MAC (f) is calculated by max-pooling over the convolutional layer activations. 
\begin{center}
    $\displaystyle  \begin{array}{{>{\displaystyle}l}}
f\ =[ f_{1} \ ,\ ...,\ f_{k} ,...f_{K}]^{T} \ with\ f_{k} \ =\ max_{x\ \epsilon \ X_{k}} x\\
\end{array}$
\end{center}
Where, $X_k$ is the $k^{th}$ feature map, of the 3D Feature Output of the Convolutional Layer. It is assumed that, 3D Output has shape of H x W x K.\\
Next, as described in \cite{tolias2015particular}, we take rectangular regions $R_i$ over the feature map, and calculate MAC for them. This gives us the RMAC Feature Matrix.
\begin{center}
    $\displaystyle  \begin{array}{{>{\displaystyle}l}}
f_{R_{i}} \ =[ f_{R_{i} ,1} \ ,\ ...,\ f_{R_{i} ,k} ,...f_{R_{i} ,K}]^{T} \\ with\ f_{R_{i} ,k} \ =\ max_{x\ \epsilon \ R_{i} ,k} x\\
\end{array}$
\end{center}
The regions have width $2*min(W, H)/(l + 1)$ and have an overlap of at-least 40 \%.Finally, we sum all regions features into a single vector.
\begin{center}
   $\displaystyle  \begin{array}{{>{\displaystyle}l}}
F_{j} \ =\ \sum ^{N}_{i\ =\ 1} f_{{R}_{i}} \ =[ \ \sum ^{N}_{i\ =\ 1} f_{R_{i} ,1} \ ,\ ...,\ \sum ^{N}_{i\ =\ 1} f_{R_{i} ,k} ,...\\
\ \ \ \ \ \ \ \ \ \ \ \ \ \ \ \ \ \ \ \ \ \ \ \ \ \ \ \ \ \ \ \ \ \ \ \ \ \ \ \ \ \ \ \ \ \ \ \ \ \ \ \ \sum ^{N}_{i\ =\ 1} f_{R_{i} ,K}]^{T} \ \\
\end{array}$
\end{center}
Here j is the layer number of convolutional feature maps, $F_j$ is RMAC features, N is the number of total regions. MS - RMAC, is then formed by calculating RMAC for multiple CNN layers, and then concatenating them.
\begin{center}
    $\displaystyle MF\ =\ [ F_{1} ,...,\ Fj,...,F_{L}]$
\end{center}
\begin{figure}[ht!]
    \centering
    \includegraphics[width = \linewidth]{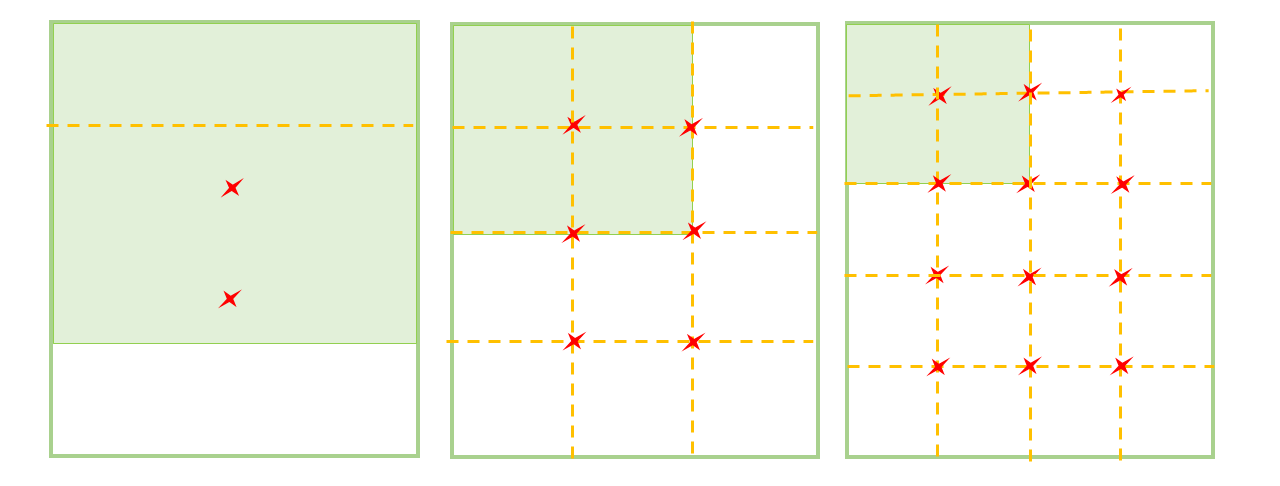}
    \caption{MS-RMAC Feature \cite{li2017ms}. This figure shows the regions extracted at 3 different scales. Green colored portion shows the currently extracted region. Centers are shown using red cross. Shaded portion slides along dashed lines in horizontal and then vertically downward directions}
    \label{fig:inceptv3}
\end{figure}

\subsubsection{}

\subsection{Retrieval Strategy}
In order to retrieve images based on content, we first calculate a representative vector for each class. This vector may be the element-wise average of global descriptors for all the images in the class or just the Global descriptor of any on particular image of the class. When an query image is given, the image is passed through the model to generate its Global descriptor. This descriptor is then compared with the representative vectors of each class. We have calculated the l2 Norm between the two vectors as a comparison. Images from the classes whose Representative vector is closer to that of the query image is returned as retrieved images. Another stage can also be added, in which the Global descriptor of the query
image is compared with the Global descriptors of images in classes found to be the closest to the query image. This can give more accurate and fine tailored results for the query.\\
 
\section{Experiments}
The Base-model chosen must be large enough to accommodate the complexity of the problem of classifying images. We have chosen InceptionV3 \cite{szegedy2016rethinking} as the base model as it has been proven to be 
capable of giving excellent performance in image classification tasks. In order to achieve significant accuracy the base model must be trained further for our specific task. This can be done for entire model mentioned in \cite{Girshick_2015_ICCV} or in the traditional way. Training of a few last layers is also found to be effective as in the case of Transfer learning. We have achieved significant accuracy by training a minimum of last 90 -100 layers.

The Branches generating the Descriptors also has to be larger than a minimum size. We have found that the performance of branches start to improve only when they are 3-4 layers deep and the hidden units layer of the the branches is comparable to the input features to the branch. These size constraints also depend on the feature and training method being used. For getting a descriptor
based on local features like \cite{opitz2017bier} ,\cite{szegedy2016rethinking}, We need a larger network.

\def\arraystretch{1.5}%
\begin{table}[htbp]
    \centering
    \caption{Studying behaviour of model on various pretrained architectures }
    \begin{tabular}{|c|c|c|}
    \hline
    Name of Architecture & Training Accuracy & Validation Accuracy \\
    \hline
    ResNet50 & 88.94\% & 74.21\%\\
    \hline
    \textbf{Inception v3} & \textbf{99.46\%} & \textbf{85.54\%} \\
    \hline
    VGG 19 & 91.23\% & 81.72\%\\
    \hline
    \end{tabular}
    \label{tab:no_layer}
\end{table} 

The size of the output from each branch affects its performance significantly. The output of branch trained by conventional method for minimizing the loss must be comparable to that of the input size of the branch. For branches trained by other methodologies, the output is dictated by the specifics of the method. But it has been found that the final model performs the best if the output from each branch  is near equal to each other as it ensures equal weightage of features extracted be each branch in the final and common Fully connected layer.
\section{Results and Discussion}

The base model fails to perform without any further training partly due to data mismatch and partly due to the fact that, most of the models are trained for image classification and object recognition tasks.
 In these tasks the difference between the samples from each class is more explicit than the difference in the samples from the classes of dataset used for Image retrieval learning. Therefore the base model must be 
fine tuned to learn finer details about the input images. Our approach does not guarantee the end-to-end training feature proposed by [multiple global descriptors], but offers more diverse ways to train.\\

\def\arraystretch{1.5}%
\begin{table}[htbp]
    \centering
    \caption{Capturing the most effective training accuracy by tuning the number of layers to be trained}
    \begin{tabular}{|c|c|c|}
    \hline
    Layers Freezed for Training & Training Accuracy & Validation Accuracy \\
    \hline
    170 & 98.89 \% & 79.24 \%\\
    \hline
    \textbf{200} & \textbf{99.46 \%} & \textbf{85.54 \%}\\
    \hline
    \end{tabular}
    \label{tab:no_layer}
\end{table}
\section{Conclusion}
In this paper, we have implemented a neural Network classifier, which has structure, similar to that of voting based system, in which the final outcome is decided based on the weights assigned to the individual outcomes of the multiple classifiers. However, the difference lies in the final stage itself. In the former case, where each of the multiple classifier is trained extensively, to perform well, here, the task of retrieval is done by using single combined feature vector, which is in turn made up by the output features of the multiple classifiers. However, since here the task is to find, most similar images, the combined feature vector can perform much better as it acts as single, flattened representative of that image. In the former case, multiple feature vector, imply multiple similarity measurements i.e more complex system. Thus the task of image retrieval is ore simplified here.

\bibliographystyle{ieeetr}
\bibliography{main.bib}

\end{document}